\pdfoutput=1

\documentclass[11pt]{article}

\usepackage[preprint]{acl}

\usepackage{times}
\usepackage{latexsym}

\usepackage[T1]{fontenc}

\usepackage[utf8]{inputenc}

\usepackage{microtype}

\usepackage{inconsolata}

\usepackage{graphicx}

\usepackage{booktabs}
\usepackage{multirow}

\usepackage{amssymb}
\usepackage{enumitem}
\usepackage{subfigure}

\newcommand{\ours}{DCMM-SQL }
\newcommand{\our}{DCMM-SQL}

\newcommand{\revised}[1]{\textcolor{black}{#1}}

\title{DCMM-SQL: Automated Data-Centric Pipeline and Multi-Model Collaboration Training for Text-to-SQL Model}

\author{Yuanzhen Xie$^{1}$ \, Liu Ye$^{1}$\, Jiqun Chu\, Mochi Gao$^{1}$\, \\ \textbf{Hehuan Liu$^{1}$\, Yunzhi Tan$^{1}$\, Bo Hu$^{1}$\, Zang Li$^{1}$} \\
$^1$Platform and Content Group, Tencent\\
\texttt{\{xieyzh3, tummasterly\}@gmail.com}}

\begin{document}
\maketitle
\begin{abstract}
Text-to-SQL tasks have gained attractive improvements since the release of ChatGPT. Among them, agent-based frameworks have been widely used in this field.
However, the impact of data-centric strategies on text-to-SQL tasks has rarely been explored. 
In this paper, we systemically design a fully automated data-centric pipeline for text-to-SQL tasks, including \emph{adaptive data repair}, which can automatically find and fix errors in the training dataset;
and \emph{error data augmentation}, where we specifically diffuse and enhance erroneous data predicted by the initially trained models.
Meanwhile, we propose a Multi-Model collaboration training schema, aiming to train multiple models with different augmented data, enabling them to possess distinct capabilities and work together to complement each other, because it has been found that the capability of a single fine-tuned model is very limited.
Furthermore, we utilize an ensemble strategy to integrate the capabilities of multiple models to solve a multiple-choice question, aiming to further improve the accuracy of text-to-SQL tasks.
The experiment results and ablation study have demonstrated the effectiveness of data-centric pipeline and Multi-Model(MM) interactive iterative strategies, achieving first place in lightweight text-to-SQL models (within 70B).

\end{abstract}

\section{Introduction}

Text-to-SQL \cite{pourreza2023dinsql, li2024can, xie2024decomposition} aims to convert natural language into SQL code to perform database operations, thereby lowering the barrier to database manipulation. With the advancement of large language models (LLMs), significant progress has been made in the field of text-to-SQL. Currently, there are three main approaches based on LLMs: 1) designing agent strategies based on general large models~\cite{pourreza2023dinsql,xie2024decomposition}; 2) fine-tuning models specifically for the text-to-SQL domain~\cite{li2024codes}; and 3) combining the capabilities of both agent strategies and model fine-tuning~\cite{gao2024xiyan}. These approaches tend to focus more on enhancing the final outcome by designing highly refined agent strategies, leading to the gradual maturation and convergence of these strategies. For instance, both Xiyan-SQL~\cite{gao2024xiyan} and CHASE-SQL~\cite{pourreza2024chase}, which are ranked top 10 in the test set of the current Bird benchmark, can be summarized as employing similar four modules: schema linking, SQL generation, SQL correction, and ensemble module.

In contrast, we observe that research on advancing text-to-SQL models through data and model training optimization strategies remains relatively scarce. 
Annotating text-to-SQL data requires significant time and expertise from professionals. Therefore, it is essential to explore a comprehensive, data-centric framework that includes synthetic data generation, augmentation, and verification schemes.
Such efforts are critical to advancing the production of text-to-SQL data and model development in the industry.

In this paper, we adopt the aforementioned agent-based framework to accomplish the text-to-SQL task, which includes schema linking, SQL generation, SQL correction, and ensemble module to accomplish the text-to-SQL task. Our work emphasizes a \textbf{D}ata-\textbf{C}entric approach and a \textbf{M}ulti-\textbf{M}odel collaboration training strategy (abbreviated as \textbf{\our}).
On the data side, based on the design of comprehensive data synthesis and augmentation as well as data validation schemes, we propose adaptive data repair and error data augmentation strategies. The former involves using a validation model to judge the results of the initial training model's inference against the standard results and then performing strategic repairs; the latter involves synthesizing and augmenting erroneous example data to obtain more similar data after validation.
To further unlock the model's potential, we introduce a two-step training strategy based on multi-model collaboration strategy in the model training phase: the first step is the initial training phase with directly supervised fine-tuning (SFT); the second step leverages the erroneous data generated from the first stage's inference for diffusion (to better learn from past mistakes) and combines it with the original data for an active learning training stage.
In the ensemble module, diverging from prior work, we maintain a data-centric focus. Recognizing that different models exhibit preferences for specific data patterns, we train multiple models on diversified datasets and employ the ensemble module to select the optimal prediction.

Our contributions are as follows:
\begin{itemize}
    \item 
    To the best of our knowledge, we are the first to design a fully automated data-centric pipeline for text-to-SQL tasks, which includes adaptive data repair, data verification, data synthesis, and data augmentation and error data augmentation to automatically find erroneous data and enhance it, thereby improving the performance of text-to-SQL models in a targeted manner.
    \item An effective two-step training mode of multi-model collaboration, including preliminary training and active learning training, is proposed. Joint data synthesis and enhancement models, data verification models, and text-to-SQL models to optimize text-to-SQL models.
    \item Good results have been achieved in the field of smaller-volume models by transforming the training process of validation and ensemble models into a multiple-choice accuracy optimization problem. Different data sets have been used to train multiple models, further enhancing the accuracy of the text-to-SQL task.
\end{itemize}
\section{Related Work}

\paragraph{Data Generation}
Along with the introduction of Large Language Models (LLMs) \cite{achiam2023gpt}, many tasks are increasingly gaining huge improvement \cite{lai2024lisa, lee2024mcs, zhang2024sentiment}. However, the performance of downstream tasks still depends on the quality of datasets \cite{li2023quantity}. As model sizes grow exponentially, large-scale and high-quality datasets are urgently required. To solve this issue, there are generally two different technologies to expand dataset using LLMs, data synthesis and data augmentation \cite{li2022data}, where the former creates entirely new samples from scratch or based on generative models, and the latter enhances existing data samples through transformations. \citet{wang2024survey} proposed an LLM-oriented data synthesis and augmentation method, emphasizing the full life cycle from data preparation to applications, with the ultimate goal of improving LLMs themselves through data-centric techniques.

In the field of text-to-SQL, which aims at converting natural language queries into SQL queries that can be executed by a database, some works \cite{li2023resdsql, wang2024macsql} focus on the Agent framework \cite{castelfranchi1998modelling, yang2023auto}, where the text-to-SQL task is broken down into several manageable smaller components, such as schema linking, SQL generation or SQL correction \cite{maamari2024death} etc. Some works utilized few-shot examples to improve the performance of SQL generation through Retrieval-Augmented Generation (RAG) method \cite{thorpe2024dubo, vichev2024ragsql, talaei2024chess}.
However, the impact of data enhancement methods on text-to-SQL tasks has rarely been comprehensively explored. In this paper, we systematically design a text-to-SQL oriented, data-centric pipeline, for two-step SFT training (see Figure \ref{fig: SQL generation module}). Through the automated data-centric pipeline (see Figure \ref{fig: SQL generation module}), we can automatically discover the error data of text-to-SQL model and repair model weaknesses through error-diffusion strategies. We have also tried different augmentation strategies and conducted multiple ablation experiments to demonstrate the effectiveness of different data-centric techniques on text-to-SQL tasks.

\paragraph{Multi-Model(MM) Collaboration}
LLMs have shown strong capabilities in many tasks. However, each LLM has its specific strengths and weaknesses. Many studies have demonstrated that combining multiple LLMs can enhance overall performance.
Such as, \citet{lu2024merge} explored different collaborative strategies of LLMs and categorized them into three approaches, including merging, ensemble and cooperation.
In the field of text-to-SQL task, \citet{gao2024xiyan} employed a multi-generator ensemble strategy to improve candidate SQL generation. \citet{pourreza2024chase} proposed CHASE-SQL, which leverages LLMs' intrinsic knowledge to generate diverse and high-quality SQL candidates using different LLM generators.
Multi-Model collaboration strategy has been used many times in our paper, where we utilize different text-to-SQL models to generate multiple SQLs and then select the best SQL generation through a selection module. We also leverage the ensemble collaborative capabilities of multi-LLMs to double check the semantic consistency of LLM-augmented query-sql pairs.

\section{Method}

\begin{figure}[ht]
  \centering
  \includegraphics[width=0.9\linewidth]{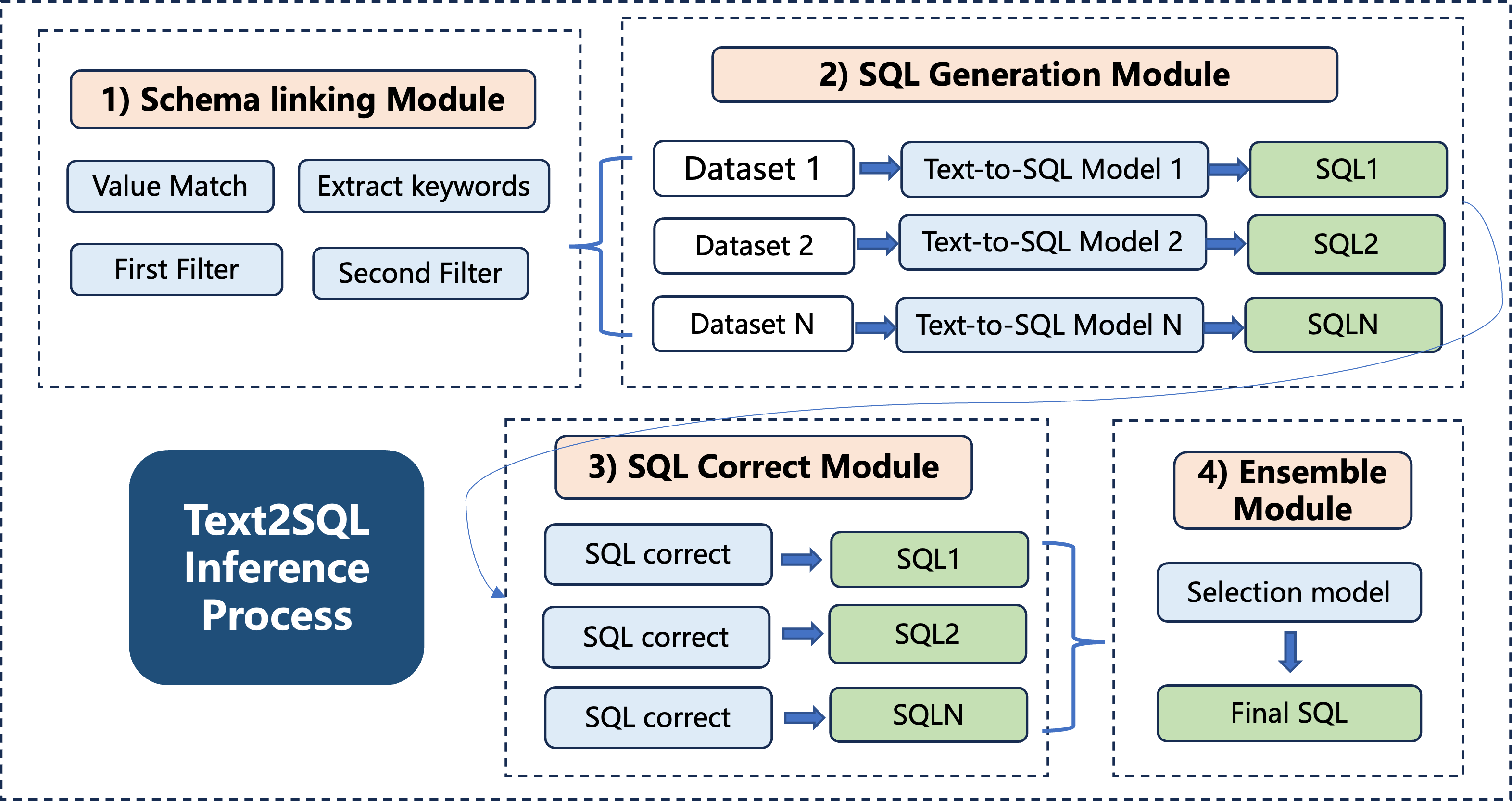}
  \caption{ \revised{The overall structure of the \ours method.}}
  \label{fig: MMSQL model}
\end{figure}

Recent researchers \cite{wretblad-etal-2024-understanding,ranaldi-etal-2024-investigating,alber2025medical} have shown that the presence of a few erroneous samples can cause a significant degradation in model performance.
In preliminary experiments, we also found that there are some erroneous samples in the training set of the text-to-SQL mainstream benchmark - Bird \cite{li2024can}, as shown in Figure \ref{fig:error_case} (See Appendix \ref{appendix:examples}).
However, the lack of high-quality data is one of the serious problems in real-world applications; meanwhile, there are relatively few studies on SQL data verification and data repair.
\revised{Therefore, we have designed an automatic data-centric solution (including \emph{Adaptive Data Repair} and \emph{Error Data Augmentation} in Figure \ref{fig: SQL generation module}) for the text-to-SQL task to find and repair some erroneous data in the training dataset. Furthermore, we have performed erroneous data diffusion and augmentation (Section \ref{subsec: Erroneous Data Enhancement}) on the data with errors in the preliminary training models, which exactly represent the weakness of the models, for the second step, active learning training.}
This automatic data-centric solution has certain task universality and can be easily adapted to other tasks.

In addition to the data-centric strategy, we also propose a multi-model interactive iterative architecture to further improve the performance of text-to-SQL models.
Unlike previous research, simply using different prompts or different temperature settings to produce multiple SQL candidates, we have designed a multi-LLMs collaboration training strategy to train multiple text-to-SQL models, where different models are trained using different erroneous augmentation data, each of these models has its own strengths, and can, to a certain extent, work together to compensate for each other's shortcomings.

The workflow of our proposed DCMM-SQL is shown in Figure \ref{fig: MMSQL model}, including Schema Linking, SQL Generation, SQL Correct, and the Ensemble module. 
First, the database information related to the problem is retrieved through the schema linking module; then multiple SQL candidates are generated through different SQL models fine-tuned with different data strategies in the SQL generation module, where the different data are diffused and enhanced through our proposed automatic data-centric automatic pipeline; SQL post-verification optimization is conducted for each SQL result in the SQL correct module; finally, multiple SQL results are given to the selection module to output the final best SQL answer.

\subsection{Schema Linking Module}
\label{subsec: Schema Linking Module}

The database may contain many tables and fields that are irrelevant to the problem, and redundant information can reduce the accuracy of the SQL generation model. Schema linking can associate keywords in natural language with elements in the database schema (such as tables, fields, field values, etc.) to filter out the most relevant information. Our schema linking solution consists of four steps: Value Match, First Filter, Extract Keywords, and Second Filter. The first step is consistent with the Codes\cite{li2024codes} method, using the BM25 algorithm and longest substring matching to obtain basic value data. In the second step, a trained BERT model is used to classify and score tables and fields, selecting the top $N$ for initial filtering while ensuring the recall rate. The third step uses the advanced large language model (LLM) to extract keywords from the question. In the final step, the LLM uses the keywords from the first step to refine the results obtained in the second step, completing the second filtering.

\subsection{SQL Generation Module}
\label{subsec: SQL Generation Module}

\begin{figure*}[ht]
  \centering
  \includegraphics[width=0.6\linewidth]{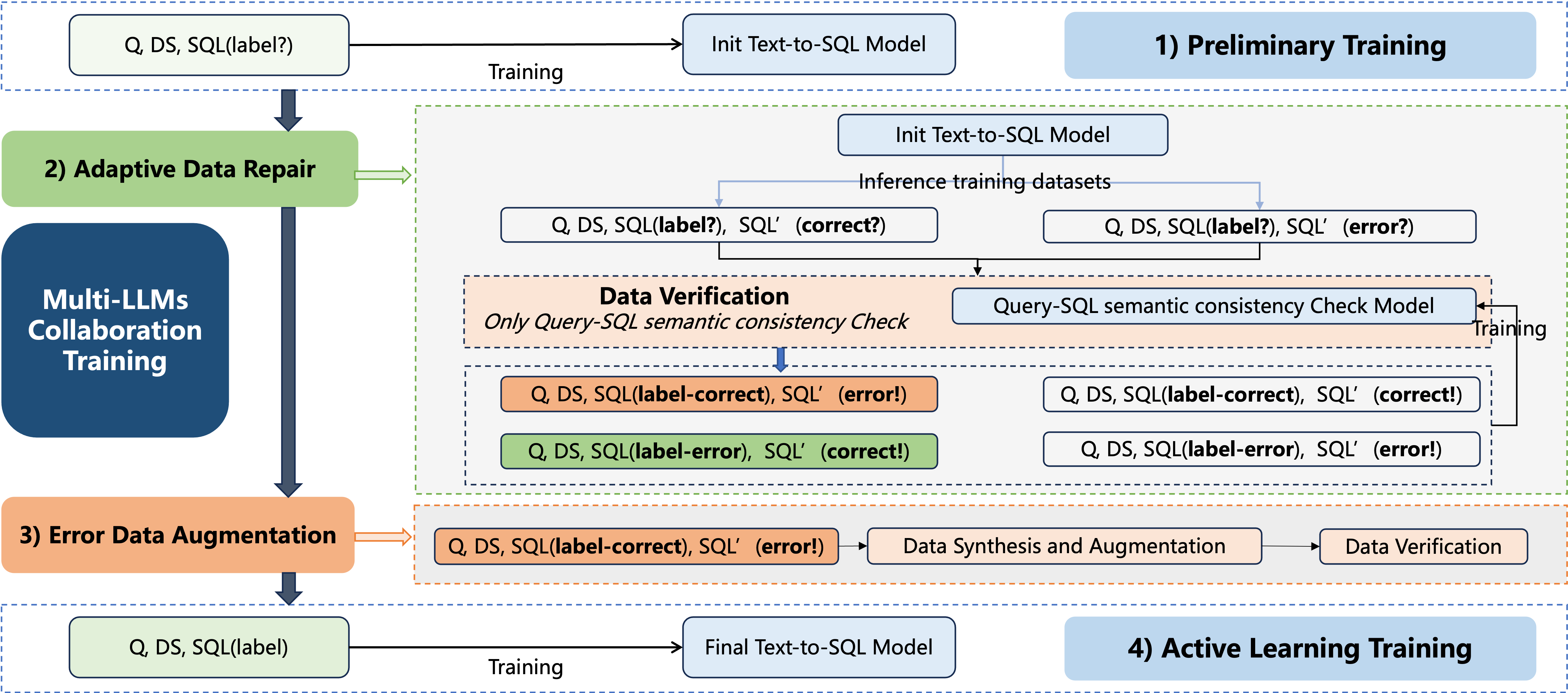}
  \caption{ \revised{The overall structure of our propsoed Multi-LLMs collaboration training consists of two steps: first, 1) preliminary training with original data, then 4) active learning training with augmented erroneous data, where erroneous training data is found and repaired in 2) adaptive data repair, and then diffused and augmented by 3) error data augmentation. ($Q$: question, $DS$: database schema, $SQL$: original annotated SQL, $SQL'$: predicted SQL by init text-to-SQL model)}
  \label{fig: SQL generation module}}
\end{figure*}

\revised{This module aims to generate accurate SQL statements. 
To generate more accurate SQL, we intend to generate multiple candidates in this module and then select the best final answer through an ensemble module, because the capabilities of a single model are limited.
Hence, a two-step training strategy including preliminary training and active learning training, as shown in Figure \ref{fig: SQL generation module}, is proposed.
Rather than directly applying different temperature settings under a single model to generate multiple SQLs, we fine-tune multiple text-to-SQL models using different data produced by a data-centric pipeline.
This data-centric pipeline (see Figure \ref{fig: SQL generation module}) can automatically discover the weaknesses of the initial text-to-SQL models and fix them by adaptive data repair, then diffuse them through error data augmentation.
Given that text-to-SQL models trained with different datasets have different preference distributions, we can combine the different features of multiple SQL models to generate more accurate SQLs.}

\subsubsection{Adaptive Data Repair}

Some erroneous data samples in the Bird training set are illustrated in Figure \ref{fig:error_case} in the Appendix \ref{appendix:examples}. Relevant studies \cite{wretblad-etal-2024-understanding, ranaldi-etal-2024-investigating, alber2025medical} have shown that these contaminated data can negatively impact the performance of text-to-SQL models. We have designed a fully automated intelligent data repair module, as shown in Figure \ref{fig: SQL generation module}, which can automatically identify and correct some of the erroneous data in the training set. Specifically, the initial SQL model is trained on the training set in the preliminary training and then used to re-predict the training data to obtain predicted $SQL'$. Utilizing the data verification module introduced below to verify $SQL$ (the original annotated SQL, which contains some erroneous samples) and $SQL'$, and the data where the validation module deems $SQL'$ correct but $SQL$ erroneous are selected for replacement.
Through the aforementioned approach, we identified and corrected 143 data instances in the Bird dataset and 95 instances in the Spider dataset.

\paragraph{Data Verification}

In terms of data verification, the following three types of verification methods are proposed to ensure the quality of synthetic data:

\begin{enumerate}
    \item \textbf{Query Check}: The model-as-predictor methods are used to perform fluency and similarity checks of synthetic queries. (In the experiment, we use the Hunyuan model.)
    
    \item \textbf{SQL Check}: The legality of SQL statements is ensured by using syntax checkers (rule-based tools, such as sqlglot\footnote{https://github.com/tobymao/sqlglot}) and execution verification (execution engines).

    \item \textbf{Query-SQL Semantic Consistency Check}: Three solutions are designed to verify whether the SQL fully addresses the content mentioned in the query: a) multiple LLM Check via prompt in zero-shot; b) a traditional classification model; c) a fine-tuned LLM checking model.
    We employ these solutions in a multi-step filtering process to ensure the effectiveness of the data verification module. 

\end{enumerate}

In query diffusion, the query check and Query-SQL Semantic Consistency Check modules are required, while in query-SQL diffusion, the SQL check and Query-SQL Semantic Consistency Check modules are needed. Relevant prompts can be found in Appendix \ref{appendix:data_verification}.

\subsubsection{Error Data Augmentation}
\label{subsec: Erroneous Data Enhancement}

We purposely synthesize and augment the training data where predicted SQLs are false
compared to the correct SQL labels, as these unpredictable data exactly represent the weaknesses of the initial text-to-SQL model.

\paragraph{Data Synthesis and Augmentation}
In the data synthesis step, we mainly design two directions to generate query-SQL data pairs: SQL2query and query2SQL.
The former uses syntax trees and regular expressions to extract SQL table names. We can get specific table schema based on the table name. We then forward the SQL and table schema to a base model to generate an interpretation of the SQL first and then summarize it into a one-sentence user query.
The latter uses a base model to generate a variety of user queries and SQLs based on specific table schema, where we propose a series of verification strategies (see the part of Data Verification) to ensure the relational pairs of synthesized query and SQL.

In terms of data augmentation strategies, we propose two methods: Query Diffusion and Example Diffusion. Query Diffusion, involves generating a variety of queries that are semantically similar but phrased differently based on the original query. The aim of this strategy is to minimize the impact of different phrasings and ensure that the core elements of the query and SQL match effectively.
Example Diffusion, is based on a given query-SQL example, which is rewritten across different databases to generate similar types of data. The purpose of this strategy is to increase the richness and diversity of the data by replicating and modifying the original example in a variety of database environments.
Both strategies aim to improve the generalization ability and accuracy of the model by expanding and enriching the training data.

\begin{figure}[ht]
  \centering
  \includegraphics[width=0.9\linewidth]{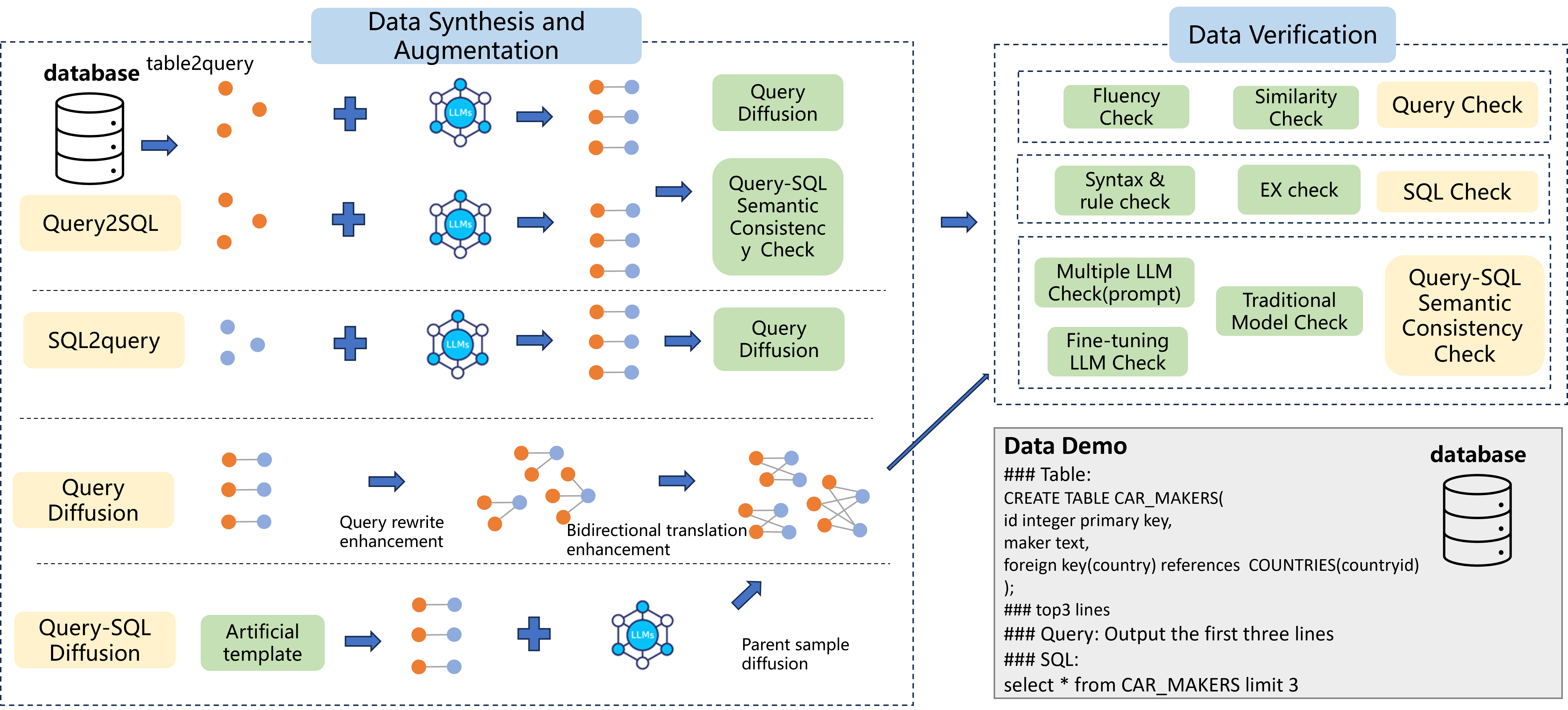}
  \caption{ The overall structure of the data synthesis, augmentation, and verification method.}
  \label{fig:data}
\end{figure}

\subsubsection{Multi-LLMs Collaboration(Training Strategy)}

To further enhance the problem-solving capabilities of text-to-SQL models, we employed interactive training with multiple LLMs and adopted a two-step SFT training approach. 
Multiple LLMs models (including GPT4o~\cite{2023gpt}, DeepSeek~\cite{guo2025deepseek}, LLaMA3~\cite{dubey2024llama}, Qwen2.5~\cite{yang2024qwen2}, Hunyuan~\cite{sun2024hunyuan}, and fine-tuned models) were used for data validation throughout the design of the interactive training. Fine-tuned models for text-to-SQL tasks and various LLMs models (Hunyuan~\cite{sun2024hunyuan}, LLaMA3~\cite{dubey2024llama}) for data generation and augmentation were also utilized.
The two-step SFT training consists of preliminary training and active learning training. In the preliminary training, original datasets are used for fine-tuning an initial text-to-SQL model. For the active learning training, the initial text-to-SQL model is used to infer the training set, identifying challenging cases, where SQLs are predicted incorrectly. These challenging cases are then enhanced with relevant data through the data synthesis and augmentation strategies mentioned earlier, with the expectation that the model will learn to solve such problems from this augmented data.

\paragraph{Preliminary Training} 

At this stage, the initial text-to-SQL model is obtained by performing SFT fine-tuning based on LoRA \cite{hu2021lora} on the base model using the training data, thus completing the preliminary training.

\paragraph{Active Learning Training}

During the active learning training process, the text-to-SQL model $M_{init}$, which has been trained in the Preliminary Training phase, is used to predict the training set. The data repair process is then applied to identify the erroneous predictions, resulting in the erroneous dataset $ED$. Subsequently, different models $M_n$ are used to perform examples and query augmentation on the dataset $ED_{error}$, yielding new datasets ($ED^{q}_{M_n}$ and $ED^{e}_{M_n}$, respectively). These datasets, along with the original training data, are combined to form new training data, which is then used to fine-tune a new model $M^*_{n}$.

\subsection{SQL Correction Module}

The primary objective of the SQL Correction module is to identify and rectify semantic and syntactic errors present in generated SQL statements. In earlier studies \cite{pourreza2024din}, the self-correction module utilized LLMs in a zero-shot setting for the purpose of self-correction, with the expectation that the LLM will correct any errors in the generated SQL statements. However, previous studies have argued that relying solely on a language model for self-correction is impractical \cite{huang2023large}. Hence, we incorporate external information into this SQL correction module to improve the correction performance.

To address two common types of errors in SQL, syntax and semantics, we propose distinct methods to provide supplementary information. For syntactic errors, where the SQL statement does not adhere to SQL syntax rules, we utilize the error information returned by the database system to achieve incremental improvements. For semantic errors, we employ a two-step correction method. The first step involves using a LLM to identify errors in SQL statements and offer modification suggestions. In the second step, the problematic SQL statements are presented to the LLM for modification, incorporating the suggestions generated in the first step. Compared to previous single-step models \cite{pourreza2024din}, the two-step method introduces additional suggestion information through the correction and discrimination capabilities of the LLM. Guided by these suggestions, the model generated in the second step can be more accurate. 
In the process, because the semantic correction strategy may still introduce grammatical errors, the semantic correction strategy is used first, and then the syntax correction strategy is used.
Please see the Appendix \ref{sec:sql_correct_prompt} for more details on prompts used in the SQL correction module.

\subsection{Ensemble Module}
In earlier approaches, the model was assigned to generate $N$ responses, followed by the application of self-consistency to enhance the accuracy of the final outcome. However, it is important to note that the most consistent answer is not always the correct one. For some questions, the model might lean towards a particular incorrect answer, leading to minimal improvement in consistency. To address this issue, the problem is reformulated as a multiple-choice question, aiming to identify the best answer using a specialized selection model.

Unlike the two-category strategy in CHASE-SQL \cite{pourreza2024chase}, our goal is to enhance the model's multi-classification capabilities for questions with an unlimited number of choices. Utilizing the table schema obtained after schema linking (subsection \ref{subsec: Schema Linking Module}) and multiple text-to-SQL models trained with diverse data, we can derive a set of candidate answers through reasoning. By grouping the execution results of these candidate answers and selecting one from each group as a candidate option, we can then fine-tune an exclusive selection model based on this data.
Please see Appendix \ref{appendix:ensemble_module} for the related prompts.

\section{Experiment}
Our experiments are designed to address the following questions:
\begin{itemize} [leftmargin=*]
    \item {\textbf{RQ1.}} How does \ours perform compared to the state-of-the-art (SOTA) baselines?
    \item {\textbf{RQ2.}} 
    What is the impact of Adaptive Data Repair and Error Data Augmentation on the text-to-SQL task, and is the data-centric pipeline scalable?
    \item {\textbf{RQ3.}} 
    How effective is each module and how do they impact the final outcome?

\end{itemize}

\subsection{Experimental Setup}

\begin{table*}[!ht]
  \caption{\revised{Comparison of Execution Accuracy on Bird and Spider Datasets Between Previous Methods and Our Method.}}
  \label{tab:overall_results}
  \centering
  \footnotesize
  \scalebox{0.85}{
  \begin{tabular}{c|ccc|ccc}
    \toprule
    Method & w/wo Training  & w/wo Data Aug &  model size & Bird Dev & Spider Dev & Spider Test  \\
    \midrule
   
    DIN-SQL + GPT-4 \cite{pourreza2023dinsql} & $\times$ & $\times$  & UNK & 50.72 & 74.2 & 85.3 \\
    DAIL-SQL + GPT-4 \cite{gao2023text} & $\times$ & $\times$ &  UNK & 54.76 &  84.4 & 86.6 \\
    MAC-SQL + GPT-4 \cite{wang2023mac} & $\times$ & $\times$  &  UNK & 57.56 & 86.75 & 82.80 \\
    \midrule
    
    ReFSQL \cite{zhang2023refsql} & \checkmark  & \checkmark & 11B & - &  \underline{88.1} & - \\

    SFT CodeS-15B \cite{li2024codes}  &  \checkmark & \checkmark   &15B & 58.47 &  84.9 & - \\

    XiYanSQL-QwenCoder-32B \cite{gao2024xiyan}  &   \checkmark & \checkmark & 32B & 67.01 & - & - \\
    OmniSQL-32B \cite{li2025omnisql}   & \checkmark  & \checkmark & 32B & 67.0 & 80.9 & 89.8 \\
    CSC-SQL \cite{sheng2025csc}   & \checkmark  & \checkmark & 32B & 71.33 & - & - \\
    XiYanSQL-QwenCoder \cite{gao2024xiyan}  &   \checkmark & \checkmark & UNK & \textbf{73.34} & - & 89.65 \\
    
    \midrule
    \ours(base qwen-32B)  & \checkmark & \checkmark  & 32B & 68.12 & 87.8 & \underline{89.75} \\
    \ours(base LLama3-70B)   & \checkmark & \checkmark  & 70B & \underline{72.69} & \textbf{88.5} & \textbf{89.84} \\
    \bottomrule
  \end{tabular}
  }
\end{table*}

\revised{Our proposed method was evaluated on two widely-used text-to-SQL benchmark datasets, Spider and Bird, with execution accuracy adopted as the performance metric. The dataset description is provided in Appendix \ref{sec:datasets}.}

\paragraph{\bf{Evaluation Metrics}}
Execution Accuracy (EX) represents the correctness of predicted SQL query execution results, which can effectively reflect the accuracy of text-to-SQL tasks. Therefore, we use EX as our sole evaluation metric.

\paragraph{\bf Baselines}

\revised{We select traditional and recently published text-to-SQL baselines for model comparison, which can be briefly categorized into two groups: (1) prompt engineering-based programs~\cite{pourreza2023dinsql,gao2023text,wang2023mac} and (2) training model-based programs~\cite{zhang2023refsql,li2024codes,gao2024xiyan,li2025omnisql}.}
Please see the Appendix~\ref{sec:baselines} for descriptions of specific related methods.

\subsection{Model Performance Comparison (RQ1)}

To validate the effectiveness of our approach, we compare it with other baselines on the mainstream text-to-SQL dataset, Bird and Spider. The overall experimental results are shown in Table \ref{tab:overall_results}. Under the execution accuracy metric, our method outperforms the listed baselines, effectively demonstrating the efficacy of our proposed data-centric pipeline and multi-model collaboration strategy in addressing complex text-to-SQL problems.

\revised{In contrast to prompt engineering-based methods~\cite{pourreza2023dinsql,gao2023text,wang2023mac}, we have achieved substantial improvements by employing a more advanced agent design and a smaller fine-tuning base, which to some extent demonstrates the effectiveness of fine-tuning models in complex tasks such as text-to-SQL.}

\revised{When compared to training model-based  programs~\cite{zhang2023refsql,li2024codes,gao2024xiyan,li2025omnisql}, the improvements in our method primarily stem from innovations in fine-tuning methods (utilizing the data-centric pipeline and multi-model collaboration training strategy).}

\subsection{Effect of Data-Centric Pipeline (RQ2)}

\begin{figure}[htbp]
  \begin{minipage}[t]{0.45\linewidth}
    \centering
    \subfigure[Data centric pipeline analysis results \label{fig:mmsql_data}]{
      \includegraphics[width=0.9\linewidth]{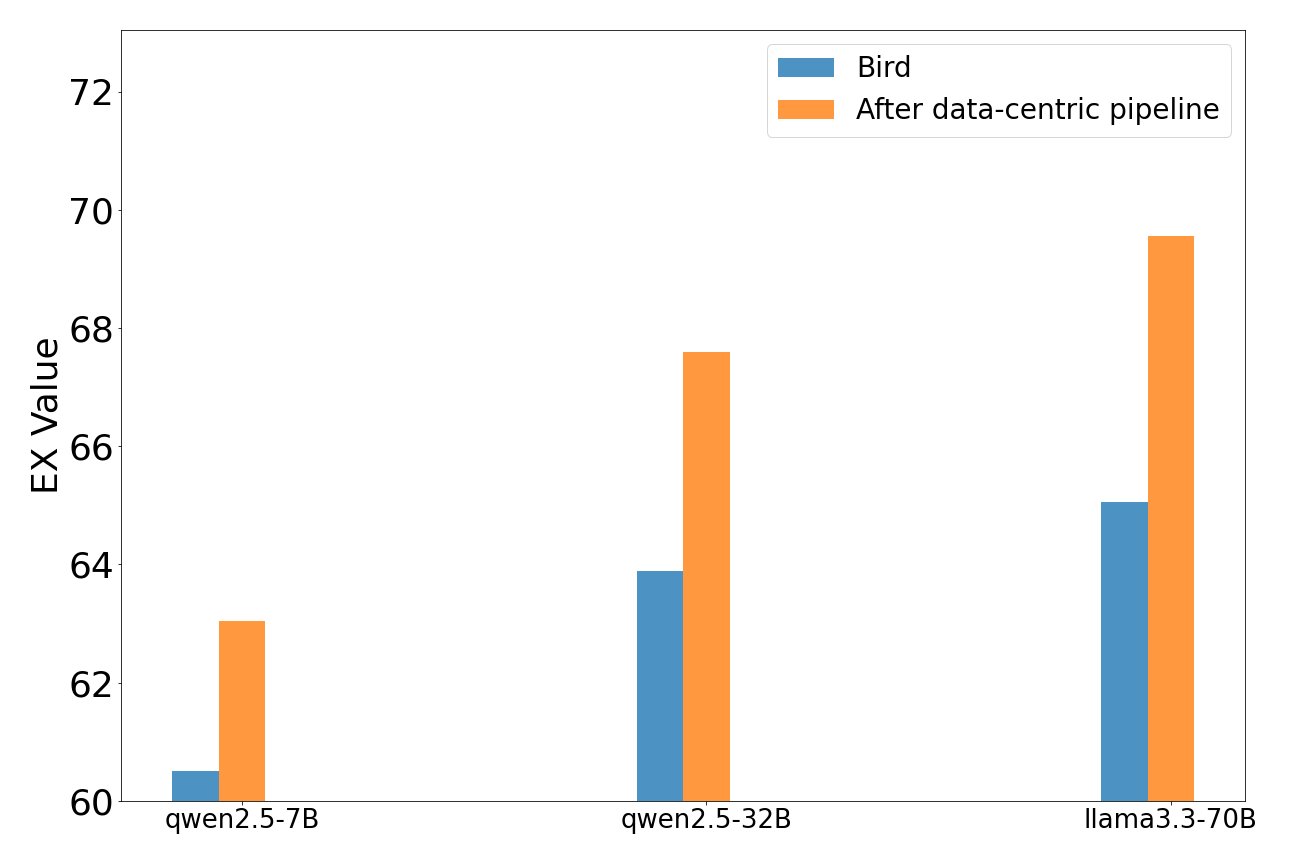}
    }
  \end{minipage}
  \centering
  \begin{minipage}[t]{0.45\linewidth}
    \centering
    \subfigure[Different datasets analysis results\label{fig:mmsql_aug}]{
      \includegraphics[width=1.0\linewidth]{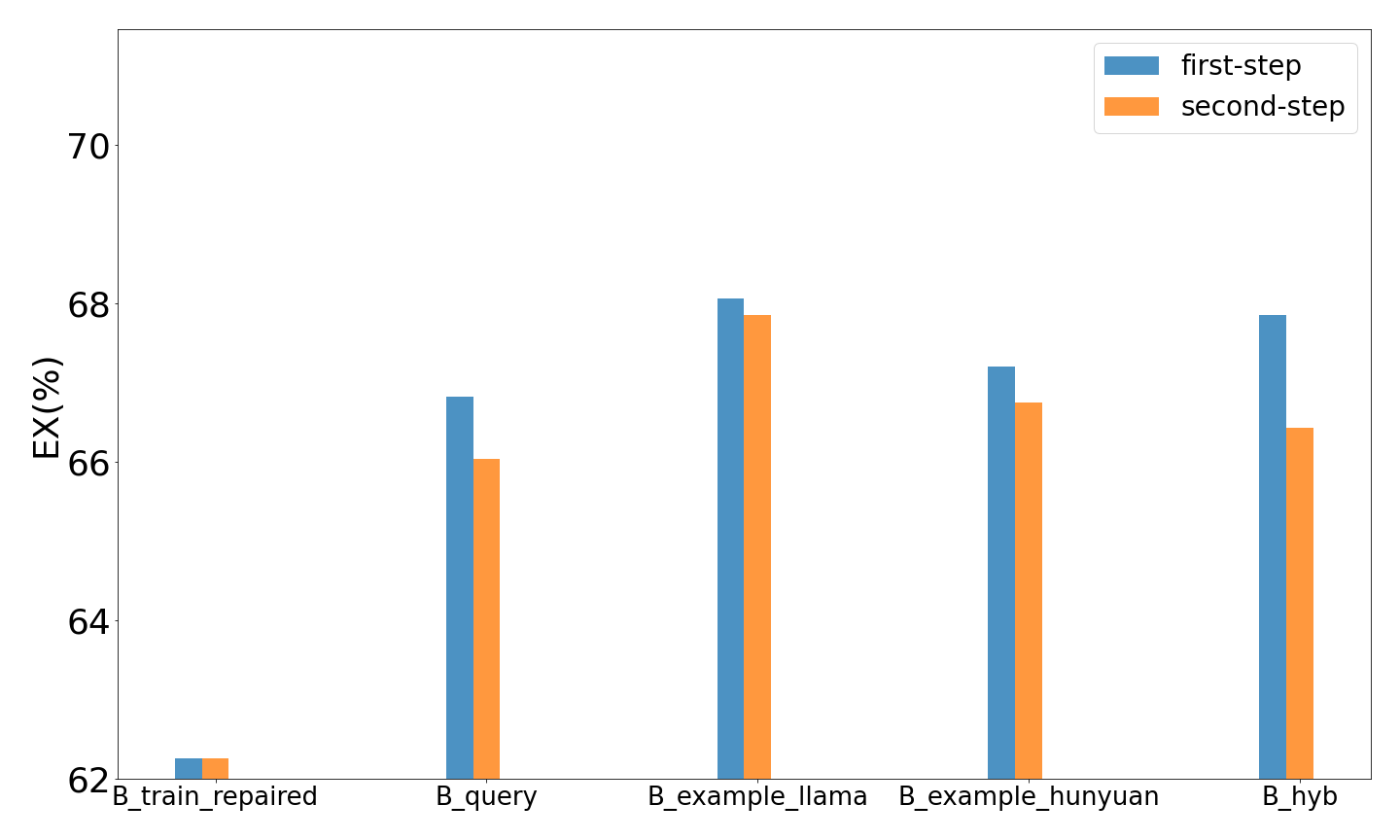}
    }
  \end{minipage}
  \caption{Data-centric pipeline analysis results.}
  \label{fig:two_images}
\end{figure}

\paragraph{Effect of Data-Centric Pipeline in Different Models}
In this section, we will delve into the role of the data-centric pipeline in the training process of large models. Specifically, this section analyzes the overall pipeline on different model bases, as well as research related to the Adaptive Data Repair and Error Data Augmentation sub-modules.

\paragraph{Effect of Adaptive Data Repair}
To investigate whether the Adaptive Data Repair strategy is effective across different model bases, we conducted experiments on the bases of qwen2.5-7B, qwen2.5-32B, and llama3.3-70B. As shown in Figure \ref{fig:mmsql_data}, the data-centric pipeline has a positive impact on different model bases. Within the same series of models, the larger the base parameters, the more significant the improvement. This indicates that the module is highly scalable and can be used in a plug-and-play manner.

\begin{table}[!ht]
    \caption{Data repaired effect on Bird Dev dataset.}
    \label{tab:data_repaired}
    \centering
    \scriptsize
    \begin{tabular}{c|c|c|c|c}
      \toprule
      Method &  simple & moderate & challenging & total \\
      \midrule
      base(7B)   &  67.46      &     51.51   &    44.83     &      60.50 \\

      repaired(7B)   &  69.51 &   53.23   &      44.83    &   62.26  \\
      \midrule
      base(32B)   & 71.14   &  53.88  &    49.66   &    63.89     \\
      repaired(32B)   & 72.11  &      56.25      &   49.66   &     65.19  \\

      \bottomrule
    \end{tabular}
  \end{table}

\paragraph{Effect of Error Data Augmentation}
We employed data augmentation and diffusion methods on the repaired Bird training set (B\_train\_repaired) to explore the impact of the Error Data Enhancement sub-module on overall performance through a two-stage data augmentation process. This section primarily investigates the following two sub-questions: 1) Is it beneficial to increase the number of iterations for training on erroneous data? 2) What are the effects of different data augmentation and diffusion methods?

First, let's introduce the data generation process and the corresponding dataset names. We performed predictions using the fine-tuned model on the B\_train\_repaired dataset, resulting in the Bird\_error dataset, which contains the prediction errors. Based on the Bird\_error dataset, we generated the first stage of augmented data using the query diffusion method: utilizing the llama3.3-70B model for query-SQL diffusion and combining it with the query-SQL Semantic Consistency Check module. This process produced the B\_example\_llama dataset based on the llama3.3-70B model and the B\_example\_hunyuan dataset based on the hunyuan model. Additionally, we constructed the B\_hyb dataset, which combines the outputs of both base models (Same with the Spider dataset). The final dataset size statistics are shown in Table \ref{tab:Bird_augmentation}.

Subsequently, the newly generated datasets are merged with the original dataset and trained together to obtain the results marked as "first-step" in the figure. The trained model is then used to infer the training set, and the incorrectly predicted data is identified for the second stage of data diffusion. This process is repeated to obtain the second set of results, as marked by "second-step" in the figure.

It can be observed from Figure \ref{fig:mmsql_aug} that among the different data strategies, B\_example\_llama performs the best. This demonstrates the effectiveness of augmenting data based on erroneous data, and that query-SQL diffusion (example) outperforms query diffusion (query). Additionally, as the number of iterations of the Error Data Augmentation sub-module increases, the final accuracy may decrease due to the introduction of excessive erroneous data. The interaction between the validation module and the data augmentation validation module deserves further exploration.

\begin{table}[!ht]
    \caption{\revised{Data Statistics of Error Data Augmentation sub-module on Bird and Spider dataset.}}
    \label{tab:Bird_augmentation}
    \centering
    \scriptsize
    \begin{tabular}{c|c|c}
      \toprule
      Datasets & first-step data count & second-step data count  \\
      \midrule
      B\_query   & 4459  &  2279   \\
      B\_example\_llama  & 3590 & 2278     \\ 
      B\_example\_hunyuan & 1845 & 2283  \\  
      B\_hyb & 9894 & 6840   \\
      \midrule
      S\_query   & 6050  &  3106   \\
      S\_example\_llama  & 605 & 570     \\ 
      S\_hyb & 6655 & 3676   \\
      \bottomrule
    \end{tabular}
  \end{table}

\subsection{Ablation Study (RQ3)}

To verify the effectiveness of each module, we conducted ablation experiments on the Bird dataset, and the results are shown in Table \ref{tab:ablation_study}. It can be observed that the schema linking, data strategy, SQL correction, and integration modules we designed brought a 1-3 percentage point improvement, highlighting the effectiveness of each module.
The experiments show that this integrated single model achieves only 69.56 on the Bird Dev dataset, significantly lower than our proposed multi-LLM collaboration method (72.69). These results are shown in Table \ref{tab:ablation_study2} demonstrate that data fusion has a positive effect on the final outcome. 

However, our integrated module still lags behind other advanced solutions. It can be observed that our solution performs better than other advanced solutions when the integrated module is removed. As shown in the table \ref{tab:ablation_study}, the results of DCMM-SQL-ensemble module are better compared to XiYanSQL-ensemble module~\cite{gao2024xiyan} and CHASE-SQL + Gemini(OS COT)-ensemble module~\cite{pourreza2024chase}. There remains considerable room for improvement in this module, which we will explore further in future research.

\begin{table}[!ht]
    \caption{Ablation study on Bird Dev dataset.}
    \label{tab:ablation_study}
    \centering
    \scriptsize
    \scalebox{0.9}{
    \begin{tabular}{c|c|c}
      \toprule
      Method &  EX(\%) & $\Delta$ EX(\%) \\
      \midrule
      \ours  & 72.69 & - \\
      \our-schema linking  &  71.06 & 1.63 $\downarrow$ \\
      \our-data aug   & 70.86 &  1.83 $\downarrow$ \\
      \our-correct sql &  71.51 & 1.18 $\downarrow$ \\
      \our-ensemble module(B\_hyb)  & 69.56 & 3.13 $\downarrow$ \\
      \midrule
      XiYanSQL-ensemble module & 68.84 & - \\
      CHASE-SQL + Gemini(OS COT)-ensemble module& 68.02 & - \\
    
      \bottomrule
    \end{tabular}}
  \end{table}

  \begin{table}[!ht]
    \caption{Ablation Study of Ensemble Modules on Bird Dev Dataset. }
    \label{tab:ablation_study2}
    \centering
    \scriptsize
    \scalebox{0.9}{
    \begin{tabular}{c|c|c}
      \toprule
      Method &  EX(\%) & $\Delta$ EX(\%) \\
      \midrule
      \ours  & 72.69 & - \\
      \our-ensemble module(B\_hyb)  & 69.56 & 3.13 $\downarrow$ \\
      \our-ensemble module(B\_query) & 67.99 & 4.7 $\downarrow$ \\
      \our-ensemble module(B\_example\_llama) & 69.17 & 3.52 $\downarrow$ \\
      \our-ensemble module(B\_example\_hunyuan) & 68.97 & 3.72 $\downarrow$ \\
      \bottomrule
    \end{tabular}}
  \end{table}

\section{Conclusion}

The agent-based framework has unlocked the immense potential of large models in the text-to-SQL task, where four relatively fixed modules (schema linking, SQL generation, SQL correction, and ensemble) are commonly formed. However, \revised{relatively few discussions focus on the research of the text-to-SQL model itself. At the same time, the preparation of the model is inseparable from data and training strategies. Our paper focuses on data and training strategies, proposing a data-centric pipeline and a Multi-LLMs collaboration training strategy based on the current mature agent solutions to further improve the accuracy of text-to-SQL tasks.}
\revised{The data-centric pipeline includes adaptive data repair and error data augmentation, where the former automatically corrects erroneous data in training, and the latter enhances error cases predicted by the initial trained models for the secondary active learning.}
\revised{The Multi-LLMs collaboration training strategy is proposed to optimize and train multiple text-to-SQL models, which are realized in an interactive iteration manner, starting from models for synthesis and augmentation, then data verification models to check the quality of synthesized data, followed with fined-tuned text-to-SQL models with augmented data, finally allowing the ensemble module to select the best output.}
\revised{Extensive experiments on the Bird and Spider datasets demonstrate that our method achieves optimal performance within a specific parameter threshold range of text-to-SQL models. }

\revised{Future work should focus on continuous optimization of three key components: (1) the schema linking module, (2) the ensemble module, and (3) the data validation model, with the ultimate goal of achieving state-of-the-art results using smaller-scale models.}

\section*{Acknowledgments}

\bibliography{main}

\appendix

\section{Appendix}
\label{sec:appendix}

\subsection{Some Examples of Incorrect Annotations}
\label{appendix:examples}
\begin{figure*}[ht]
  \centering
  \includegraphics[width=0.9\linewidth]{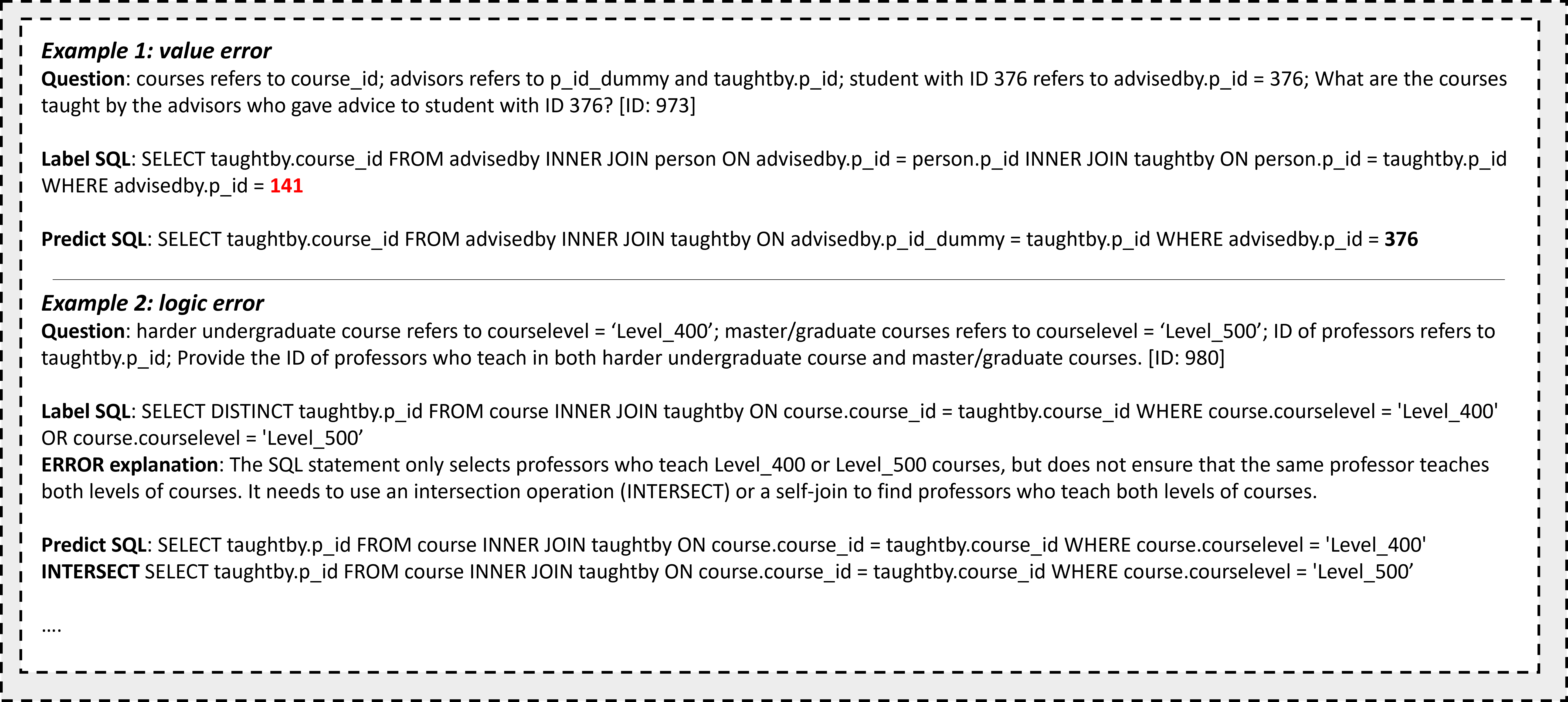}
  \caption{Some examples of incorrect annotations in the Bird training set.}
  \label{fig:error_case}
\end{figure*}

There are some wrong annotation examples in Bird's training dataset, as shown in Figure \ref{fig:error_case}.

\subsection{Datasets}
\label{sec:datasets}
\paragraph{\bf{Bird Dataset}~\cite{li2024can}} Bird is a detailed dataset with $12,751$ unique question-SQL pairs across $95$ large databases, totaling $33.4$ GB. It spans over $37$ professional fields, including blockchain, hockey, healthcare, and education. To aid models in generating accurate SQL queries, Bird incorporates external knowledge from four sources: numerical reasoning, domain-specific knowledge, synonyms, and value illustration.

\paragraph{Spider Dataset~\cite{yu2018spider}.}
\revised{The Spider dataset is a large-scale benchmark for complex, cross-domain semantic parsing and text-to-SQL tasks, comprising 10,181 questions paired with 5,693 unique SQL queries across 200 databases (multiple tables, 138 domains). The data is split into 8,659 training examples, 1,034 development examples (Spider-dev), and 2,146 test examples (Spider-test), ensuring robust evaluation of generalization in schema-rich, multilingual environments.}

\subsubsection{Regarding the Evaluation on the Spider2.0 Dataset}

We fully acknowledge the importance of Spider2.0~\cite{lei2024spider} as an emerging benchmark. During the experimental design phase, we decided not to include it in our evaluation based on the following considerations:

(1) Spider2.0~\cite{lei2024spider} currently offers only a limited number of test samples, and its training set has not been fully released. This restricts our ability to conduct model training and validation on this dataset;

(2) There are still relatively few baseline models reporting results on this dataset, making it difficult to conduct fair and systematic comparisons.

The core contribution of this work lies in proposing a general strategy for optimizing training data and enabling collaborative model training. Therefore, we selected the widely used Bird and Spider benchmarks, which are more aligned with the research theme of data-centric training. For the sake of relevance and rigor, we will cite relevant literature on Spider2.0~\cite{lei2024spider} in the camera-ready version and provide a detailed explanation in the appendix regarding the reasons for not conducting experimental comparisons on Spider2.0~\cite{lei2024spider}.

\subsection{Baselines}
\label{sec:baselines}
\begin{itemize}
\item {\textbf{DIN-SQL + GPT-4}} DIN-SQL~\cite{pourreza2023dinsql} utilizes multi-step decomposition and self-correction to effectively improve the performance of context-based learning methods on text-to-SQL tasks.
\item {\textbf{DAIL-SQL + GPT-4 + SC }} DAIL-SQL~\cite{gao2023text} systematically and extensively compares existing prompting strategies and proposes a novel integrated solution. It is the best method based on LLMs.
\item {\textbf{MAC-SQL }} MAC-SQL~\cite{wang2023mac} presents an LLM-based multi-agent framework for Text-to-SQL tasks with three agents: the Selector (condenses databases and retains relevant schemas), the Decomposer (simplifies complex questions), and the Refiner (validates and refines SQL).
\revised{\item {\textbf{ReFSQL }} ReFSQL \cite{zhang2023refsql} proposes a retrieval-augmented framework that enhances Text-to-SQL generation by incorporating structure-specific knowledge through unsupervised retrieval and contrastive learning, improving accuracy and robustness.}
\item {\textbf{SFT CodeS-15B }}  CodeS~\cite{li2024codes} is an open-source language model designed for text-to-SQL tasks, achieving high accuracy with smaller parameter sizes. It uses incremental pre-training and bi-directional data augmentation, excelling in multiple benchmarks.
\item {\textbf{XiYanSQL-QwenCoder-32B}}  XiYan-SQL~\cite{gao2024xiyan} enhances natural language to SQL tasks using a multi-generator ensemble and M-Schema for better database understanding, achieving state-of-the-art accuracy on multiple benchmarks. It combines in-context learning with supervised fine-tuning to refine SQL query candidates.
\revised{\item {\textbf{OmniSQL-32B}} OmniSQL-32B~\cite{li2025omnisql} proposes SynSQL-2.5M, a scalable framework for automatically synthesizing large-scale text-to-SQL datasets, and introduces OmniSQL, an open-source model trained on this data that achieves state-of-the-art performance. The solution addresses limitations of existing methods by generating diverse synthetic data and creating powerful yet compact models that rival larger proprietary systems.}
\item {\textbf{CHASE-SQL + Gemini}} CHASE-SQL~\cite{pourreza2024chase} is a new framework for improving Text-to-SQL tasks by using multi-agent modeling and innovative strategies to generate and select high-quality SQL candidates. It achieves state-of-the-art execution accuracy on the BIRD Text-to-SQL dataset, outperforming previous methods.
\end{itemize}

\subsection{Experiment Setup}

\textbf{Training Details of the Selection Model}: We fine-tuned the answer selection model based on the Qwen2.5-14B architecture using LoRA. Key hyperparameters were set as follows: batch size = 1, learning rate = 0.0001, number of training epochs = 1, LoRA rank = 16, optimizer = AdamW, and learning rate scheduler = cosine\_with\_restarts. The training data comprised 17,988 samples derived from the Bird dataset through reconstruction.

\textbf{Training Details of the Text-to-SQL Model}: For the text-to-SQL model, we performed LoRA-based fine-tuning using one of the following base models: Qwen2.5-Coder-7B, Qwen2.5-Coder-32B, or LLaMA3-70B. The same set of key hyperparameters was used: batch size = 1, learning rate = 0.0001, training epochs = 1, LoRA rank = 16, optimizer = AdamW, and scheduler = cosine\_with\_restarts.

\revised{\subsection{Complementary Experimental Results}
Our approach has not yet achieved state-of-the-art performance on the Bird benchmark, as the top-performing methods all employ significantly larger models. However, we have obtained competitive results in the small-model category. To achieve optimal performance in the text-to-SQL task, every module must maintain exceptionally high accuracy. Currently, our schema linking and ensemble modules show room for improvement (87\% accuracy), though this falls outside the scope of our current research. As shown in Table \ref{tab:ablation_study}, our method outperforms comparable approaches when the ensemble module is removed, demonstrating the effectiveness of our data-centric strategy to some extent.}

The CodeS\cite{li2024codes} work was published earlier and its performance is relatively weaker. As shown in Table \ref{tab:data_repaired}, our 7B model using only data repair (62.26) already outperforms CodeS-15B (58.47).  Recent state-of-the-art methods are mostly based on 32B+ models. To facilitate comparison with the best available results, we primarily selected 32B+ base models for the comparisons in Table \ref{tab:overall_results}.

\textbf{Regarding the analysis of performance degradation with increased iterations of erroneous data augmentation:}  The accuracy bottleneck of the data validation module is a critical factor.  In a manual review of 50 randomly selected samples, we identified 2 errors (including redundant column names and enumeration errors), indicating that automatic validation still contains non-negligible inaccuracies. These errors may accumulate over multiple iterations, potentially introducing additional noise.

\subsection{Prompt of Data Verification}
\label{appendix:data_verification}

\revised{\subsubsection{Query fluency Check}
Please determine whether the following statements are fluent. Answer yes or no. [query] \\}

\revised{\subsubsection{Query Similarity Check}
Please determine whether the core expressions of the following two sentences are consistent. Answer yes or no. \\
1. [query1] \\
2. [query2] \\}

\subsubsection{Multi-LLMs Check}
\#\#\# Question: [query] \\
\\
\#\#\# Database schema: [table\_info] \\ 
\\
\#\#\# SQL: [SQL] \\ 
\\
Please judge whether the SQL has completely completed the question asked in the Question. The output format is: \\ \{"Completed": "Yes or No", "Reason": "xx"\} \\

\subsubsection{Fine-tuning LLM Check}
\#\#\# Instruction:\\Based on the following table information, question, please judge whether the given SQL is correct. If it is correct, output 1, if it is not correct, output 0.\\
\\
\#\#\# Question: [query] \\
\\
\#\#\# Database schema: [table\_info] \\
\\
\#\#\# SQL: [sql] \\ 
\\
\#\#\#  Answer:

\subsection{Prompt of SQL Correct Module}
\label{sec:sql_correct_prompt}
\subsubsection{The First Step of Semantic Correction Strategy}
\#\#\# Instructions:\\
You are a SQL expert, and you are now checking a SQL statement that may have an error. Your corrections follow the principle of minimal modification, that is, if you are not sure whether a modification is necessary, you will not modify it. You only need to consider the following steps, You can't include other columns. Let's check the SQL statement in the following aspects step by step:\\
1. Restate the problem requirements and the evidences at the beginning of the question, and clarify the required output and conditions. \\
2. Let's parse SQL query, find out what columns it selects and find out what the condition is.\\
3. Let's determine whether there is a better alternative to the columns in the tables to replace the columns selected in the SQL query. You can't include other columns, only judge whether you need to replace the original columns. Here is an example:\\
\#\#\# Questions: In which mailing street address can you find the school that has the lowest average score in reading? Also give the school's name.\\
\#\#\# SQL Query: SELECT schools.mailstrabr, schools.school FROM satscores INNER JOIN schools ON satscores.cds = schools.cdscode ORDER BY satscores.avgscrread LIMIT 1\\
\#\#\# Tables: CREATE TABLE schools (\\
...\\
"mailstrabr" TEXT COMMENT mailing street address; VALUES: [313 West Winton Ave.],\\
"mailstreet" TEXT COMMENT mailstreet; VALUES: [313 West Winton Avenue],\\
...\\
)\\
\#\#\# Response:The question asked which mailing street address. In general, the default answer here is the full name of the street without abbreviation, that is, schools.mailstreet, not schools.mailstrabr. So the final output column should be schools.MailStreet, schools.School.\\
4. Let's determine whether the conditions for the join operation are reasonable, including whether the condition using a foreign key? Here is an example:\\
\#\#\# Questions:\\
Give the names of the schools with the percent eligible for free meals in K-12 is more than 0.1 and test takers whose test score is greater than or equal to 1500?\\
\#\#\# SQL Query:\\
SELECT DISTINCT frpm.`school name` FROM frpm INNER JOIN satscores ON satscores.sname = frpm.`school name` WHERE frpm.`percent (\%) eligible free (k-12)` > 0.1 AND satscores.numge1500 > 0\\
\#\#\# Tables:\\
Foreign\_key: [frpm.cdscode = schools.cdscode, satscores.cds = schools.cdscode]\\
\#\#\# Response: The join condition used in the sqlquery is satscores.sname = frpm.`school name`. Neither of these two columns is a Foreign\_key. The correct condition should be satscores.cds = frpm.cdscode.\\
5. Let's refer to the evidences in step1 and determine whether the conditional judgment matches the requirements of step 1, including whether the column selected by the condition is correct and whether the enumeration value selected by the condition is correct. Here is an example:\\
\#\#\# Questions: Of all the schools with a mailing state address in California, how many are active in San Joaquin?\\
\#\#\# SQL Query: SELECT count(cdscode) FROM schools WHERE mailstate = 'CA' AND statustype = 'Active' AND county = 'San Joaquin'\\
\#\#\# Response: The condition in the question is wrong. From the question, we can infer that San Joaquin is a city, not a county. So the condition should be city = ‘San Joaquin’. \\
6. Let's determine if there is an error in the calculation formula. You only need to determine whether the formlula of the SQL statement is consistent with the expression in the question, and do not need to consider other issues. You can't include other columns. Here is an example:
\#\#\# Questions: percent (\%) eligible frpm (ages 5-17) can be acquired by `free meal count (ages 5-17)` / `enrollment (ages 5-17)` * 100\%; Which schools served a grade span of Kindergarten to 9th grade in the county of Los Angeles and what is its Percent (\%) Eligible FRPM (Ages 5-17)?\\
\#\#\# SQL Query: SELECT frpm.`percent (\%) eligible frpm (ages 5-17)` FROM frpm INNER JOIN schools ON frpm.cdscode = schools.cdscode WHERE schools.county = 'Los Angeles' AND schools.gsserved = 'K-9'\\
\#\#\# Tables: CREATE TABLE frpm (\\
"enrollment (ages 5-17)" REAL COMMENT enrollment (ages 5-17); VALUES: [1070.0,376.0],\\
"free meal count (ages 5-17)" REAL COMMENT free meal count (ages 5-17); VALUES: [553.0,182.0],\\
"percent (\%) eligible frpm (ages 5-17)" REAL COMMENT percent (\%) eligible frpm (ages 5-17); VALUES: [0.65607476635514,0.484042553191489],\\
PRIMARY KEY ("cdscode")\\
)\\
\#\#\# Response:The calculation formula in the question requires `free meal count (ages 5-17)` / `enrollment (ages 5-17)` * 100\%. So we cannot directly use `percent (\%) eligible frpm (ages 5-17)`, but instead use frpm.`FRPM Count (Ages 5-17)` * 100 / frpm.`Enrollment (Ages 5-17)`\\
\\
7. Please summarize the output of steps 3, 4, 5, and 6 and determine whether the SQL statement matches the requirements of the problem. If steps 3, 4, and 5 do not indicate that there is a problem with the SQL query, you should reply ‘The SQL query is correct, and no modifications are needed.’\\
\\
\#\#\# Questions:
 [question] \\
\#\#\# SQLite Query: 
 [SQL] \\
\#\#\# Tables:
 [table\_info] \\
\#\#\# Response:

\subsubsection{The Second Step of Semantic Correction Strategy}
\#\#\# Instruction: \\
For a given problem, use the provided Tables to fix any problem in the buggy SQLite query.  \\
You should pay attention to the following suggestions. \\
Your response should only include a SQLite statement. \\ 
If there are any problems, return the SQLite statement with the correct modifications. \\
If there are no problems, return the SQLite statement unchanged. \\
\\
\#\#\# Suggestions: 
[suggestion] \\
\\
\#\#\# Questions:
[question] \\
\\
\#\#\# Buggy SQLite QUERY:
[SQL] \\
\\
\#\#\# Tables:
[table\_info] \\
\\
\#\#\# FIXED SQLite QUERY:

\subsubsection{Syntax Correction Strategy}
\#\#\# Instruction:  \\
You are a database administrator and aim to correct the error in the SQL statement. The following is the error message, please correct the error so that the SQL statement can run. Your response should only include a SQLite statement.  \\
\#\#\# Error Message:
[error]   \\
\#\#\# Questions:
[question]  \\
\#\#\# Buggy SQLite QUERY: 
[sql] \\
\#\#\# Tables:
[table\_info] \\
\#\#\# FIXED SQLite QUERY:

\subsection{Prompt Example of Ensemble Module}
\label{appendix:ensemble_module}
\#\#\# Instruction: Based on the following table information and query, select the most relevant SQL. \\
\\
\#\#\# Question: \\
released in the year 1945 refers to movie\_release\_year = 1945; Name movie titles released in year 1945. Sort the listing by the descending order of movie popularity.\\
\\
\#\#\# Database:\\
CREATE TABLE movies (\\
"movie\_release\_year" INTEGER COMMENT movie\_release\_year; VALUES: [2007,2006],\\
"movie\_title" TEXT COMMENT movie\_title; VALUES: [La Antena,Elementary Particles],\\
"movie\_popularity" INTEGER COMMENT movie\_popularity; VALUES: [105,23],
"movie\_id" INTEGER COMMENT movie\_id; VALUES: [1,2],\\
"movie\_title\_language" TEXT COMMENT movie\_title\_language; VALUES: [en],
"director\_name" TEXT COMMENT director\_name; VALUES: [Esteban Sapir,Oskar Roehler],\\
"movie\_url" TEXT COMMENT movie\_url; VALUES: [],\\
"movie\_image\_url" TEXT COMMENT movie\_image\_url; VALUES: [],\\
"director\_id" TEXT COMMENT director\_id; VALUES: [131,73],\\
"director\_url" TEXT COMMENT director\_url; VALUES: []\\
PRIMARY KEY ("movie\_id")\\
)\\
\\
CREATE TABLE ratings\_users (\\
"user\_id" INTEGER COMMENT user\_id; VALUES: [41579158,68654088],\\
"user\_subscriber" INTEGER COMMENT user\_subscriber; VALUES: [0,1],\\
"user\_trialist" INTEGER COMMENT user\_trialist; VALUES: [0,1],\\
"user\_has\_payment\_method" INTEGER COMMENT user\_has\_payment\_method; VALUES: [0,1],\\
"rating\_date\_utc" TEXT COMMENT rating\_date\_utc; VALUES: [2017-06-10,2012-10-02],\\
"user\_cover\_image\_url" TEXT COMMENT user\_cover\_image\_url; VALUES: [],\\
"user\_eligible\_for\_trial" INTEGER COMMENT user\_eligible\_for\_trial; VALUES: [1,0],\\
"user\_avatar\_image\_url" TEXT COMMENT user\_avatar\_image\_url; VALUES: []\\
PRIMARY KEY ("")\\
)\\
\\
CREATE TABLE lists\_users (\\
"list\_id" INTEGER COMMENT list\_id; VALUES: [192287,192313],\\
"user\_id" INTEGER COMMENT user\_id; VALUES: [2385,15264],\\
"user\_trialist" INTEGER COMMENT user\_trialist; VALUES: [1,0],\\
"user\_has\_payment\_method" TEXT COMMENT user\_has\_payment\_method; VALUES: [1,0],v
"user\_subscriber" INTEGER COMMENT user\_subscriber; VALUES: [1,0],\\
"user\_eligible\_for\_trial" TEXT COMMENT user\_eligible\_for\_trial; VALUES: [0,1],\\
"user\_cover\_image\_url" TEXT COMMENT user\_cover\_image\_url; VALUES: [],\\
"user\_avatar\_image\_url" TEXT COMMENT user\_avatar\_image\_url; VALUES: [],\\
"list\_creation\_date\_utc" TEXT COMMENT list\_creation\_date\_utc; VALUES: [2009-12-18,2010-01-30],\\
"list\_update\_date\_utc" TEXT COMMENT list\_update\_date\_utc; VALUES: [2019-11-26,2020-05-01]\\
PRIMARY KEY ("list\_id, user\_id")\\
)\\
\\
CREATE TABLE lists (\\
"list\_title" TEXT COMMENT list\_title; VALUES: [Headscratchers],\\
"list\_movie\_number" INTEGER COMMENT list\_movie\_number; VALUES: [5,3],\\
"list\_id" INTEGER COMMENT list\_id; VALUES: [1,2],\\
"user\_id" INTEGER COMMENT user\_id; VALUES: [88260493,45204418],\\
"list\_description" TEXT COMMENT list\_description; VALUES: [],\\
"list\_comments" INTEGER COMMENT list\_comments; VALUES: [3,2],\\
"list\_url" TEXT COMMENT list\_url; VALUES: [],\\
"list\_followers" INTEGER COMMENT list\_followers; VALUES: [5,1],\\
"list\_third\_image\_url" TEXT COMMENT list\_third\_image\_url; VALUES: [],\\
"list\_second\_image\_url" TEXT COMMENT list\_second\_image\_url; VALUES: []\\
PRIMARY KEY ("list\_id")\\
)\\
\\
CREATE TABLE ratings (\\
"movie\_id" INTEGER COMMENT movie\_id; VALUES: [1066,1067],\\
"rating\_id" INTEGER COMMENT rating\_id; VALUES: [15610495,10704606],\\
"critic" TEXT COMMENT critic; VALUES: [],\\
"user\_id" INTEGER COMMENT user\_id; VALUES: [41579158,85981819],\\
"rating\_score" INTEGER COMMENT rating\_score; VALUES: [3,2],\\
"critic\_comments" INTEGER COMMENT critic\_comments; VALUES: [0,2],\\
"critic\_likes" INTEGER COMMENT critic\_likes; VALUES: [0,1],\\
"rating\_url" TEXT COMMENT rating\_url; VALUES: [],\\
"user\_trialist" INTEGER COMMENT user\_trialist; VALUES: [0,1],\\
"user\_subscriber" INTEGER COMMENT user\_subscriber; VALUES: [0,1]\\
PRIMARY KEY ("")\\
)\\
Foreign\_key: [lists.user\_id = lists\_users.user\_id, ratings\_users.user\_id = lists\_users.user\_id, lists\_users.user\_id = lists.user\_id, lists\_users.list\_id = lists.list\_id, ratings.user\_id = ratings\_users.user\_id, ratings.rating\_id = ratings.rating\_id, ratings.user\_id = lists\_users.user\_id, ratings.movie\_id = movies.movie\_id]\\
\\
\#\#\# Match value:matched contents:\\
method.summary ( Descending order , Descending Order , release , Release )\\
actor.actorid ( 1945 )\\
airlines.arr\_time ( 1945 )\\
airlines.crs\_dep\_time ( 1945 )\\
airlines.dep\_time ( 1945 )\\
airlines.op\_carrier\_fl\_num ( 1945 )\\\\
alias.zip\_code ( 1945 )\\
answer.userid ( 1945 )\\
area\_code.zip\_code ( 1945 )\\
author.author\_id ( 1945 )\\
avoid.zip\_code ( 1945 )\\
methodparameter.name ( descending , popularity , Referes )\\
keyword.keyword\_name ( popularity )\\
keyword.keyword ( popularity )\\
\\
\#\#\#SQL:\\
\\
\#\#\# Options: \\
A: SELECT movie\_title FROM movies WHERE movie\_release\_year = 1945 ORDER BY movie\_popularity DESC LIMIT 1\\
Execution result(top 10 lines): [('Brief Encounter',)])]\\
B: SELECT movie\_title FROM movies WHERE movie\_release\_year = 1945 ORDER BY movie\_popularity DESC\\
Execution result(top 10 lines): [('Brief Encounter',), ('Children of Paradise',), ('Rome, Open City',), ('Scarlet Street',), ('The Lost Weekend',), ('Spellbound',), ('Detour',), ('Mildred Pierce',), ('I Know Where I’m Going!',), ('Leave Her to Heaven',)]

\subsection{\revised{The Cost Analysis}}
\revised{The training time and GPU memory usage are challenging to compare directly with other methods. Regarding the training cost, taking the 70B model as an example, we used LORA training with two 80G A100 GPUs. Assuming that the cost of a single training session is equivalent, when compared with the resource expenditure of a single training session for an individual equivalent model, our approach is approximately equivalent to the cost of six training sessions plus the training cost of one integrated model. }

\end{document}